\documentclass[conference]{IEEEtran}
\IEEEoverridecommandlockouts
\usepackage{cite}
\usepackage{amsmath,amssymb,amsfonts}
\usepackage{algorithmic}
\usepackage{graphicx}
\usepackage{textcomp}
\usepackage{xcolor}
\usepackage{multirow}
\usepackage{makecell, multirow}
\newcommand{\fabox}[4]{\makecell{$#1$\;\scriptsize (#2$|$#3$|$#4)}}

\def\BibTeX{{\rm B\kern-.05em{\sc i\kern-.025em b}\kern-.08em
    T\kern-.1667em\lower.7ex\hbox{E}\kern-.125emX}}
\begin{document}

\title{Dual-stage and Lightweight Patient Chart Summarization for Emergency Physicians}
\author{
  \IEEEauthorblockN{
    Jiajun Wu\IEEEauthorrefmark{1},
    Swaleh Zaidi\IEEEauthorrefmark{1},
    Braden Teitge\IEEEauthorrefmark{4},
    Henry Leung\IEEEauthorrefmark{1},
    Jiayu Zhou\IEEEauthorrefmark{3},
    Jessalyn Holodinsky\IEEEauthorrefmark{2},
    Steve Drew\IEEEauthorrefmark{1}
  }
  \IEEEauthorblockA{\IEEEauthorrefmark{1}Department of Electrical and Software Engineering, University of Calgary, Calgary, AB, Canada\\
  \IEEEauthorblockA{\IEEEauthorrefmark{2}Department of Emergency Medicine, University of Calgary, Calgary, AB, Canada} 
  \IEEEauthorblockA{\IEEEauthorrefmark{3}School of Information, University of Michigan, Ann Arbor, MI, USA}    
  \IEEEauthorblockA{\IEEEauthorrefmark{4}Rockview General Hospital, Calgary, AB, Canada} 
  \{jiajun.wu1, swaleh.zaidi, leungh, jessalyn.holodinsky, steve.drew\}@ucalgary.ca,\\
  jiayuz@umich.edu, braden.teitge@albertahealthservices.ca}
}

\maketitle

\begin{abstract}
Electronic health records (EHRs) contain extensive unstructured clinical data that can overwhelm emergency physicians trying to identify critical information. We present a two-stage summarization system that runs entirely on embedded devices, enabling offline clinical summarization while preserving patient privacy. In our approach, a dual-device architecture first retrieves relevant patient record sections using the Jetson Nano-R (Retrieve), then generates a structured summary on another Jetson Nano-S (Summarize), communicating via a lightweight socket link. The summarization output is two-fold: (1) a fixed-format list of critical findings, and (2) a context-specific narrative focused on the clinician’s query.
The retrieval stage uses locally stored EHRs, splits long notes into semantically coherent sections, and searches for the most relevant sections per query. The generation stage uses a locally hosted small language model (SLM) to produce the summary from the retrieved text, operating within the constraints of two NVIDIA Jetson devices. We first benchmarked six open-source SLMs under 7B parameters to identify viable models.
We incorporated an LLM-as-Judge evaluation mechanism to assess summary quality in terms of factual accuracy, completeness, and clarity. Preliminary results on MIMIC-IV and de-identified real EHRs demonstrate that our fully offline system can effectively produce useful summaries in under 30 seconds.
\end{abstract}


\section{Introduction}

Emergency physicians often face the daunting task of manually reviewing patients' electronic health records (EHRs) under tight time constraints. 
EHRs often contain large amounts of unstructured data, making it difficult for physicians to review and locate critical information~\cite{ehrenstein2019obtaining}.
In many cases, especially for elderly patients who have difficulty communicating and have multiple prior visits, ED physicians have to spend a considerable amount of time digging through the patient's EHRs manually.
In such situations, automated summarization of patient EHRs may greatly benefit physicians by highlighting key information and providing context-specific insights, resulting in a faster and more accurate decision-making process~\cite{ahsan2024retrieving}.

\begin{figure}[t]
  \centering
\includegraphics[width=0.49\textwidth]{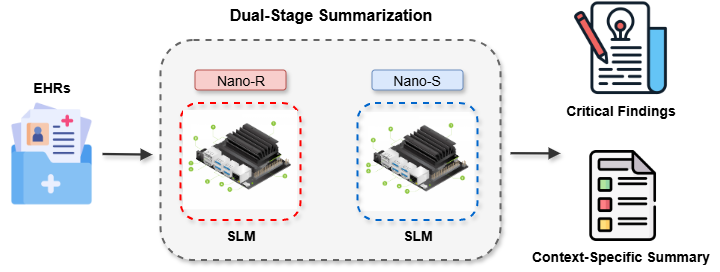}
  \caption{Dual-stage on-device architecture, enabling low-latency and privacy-preserving inference at the point of care.}
  \label{fig:architecture}
\end{figure}

Recent advances in large language models (LLMs) have demonstrated the ability to generate fluent summaries of text, suggesting a potential solution for automatic patient chart summarization~\cite{hartman2024developing, goodman2024ai, landman2024artificial}. For instance, Hartman et al.~\cite {hartman2024developing} fine-tuned an LLM to generate structured emergency department (ED) handoff notes and found that the AI-generated summaries were almost as useful and safe as physician-written ones. 
Williams et al. evaluated LLMs for drafting ED encounter summaries, demonstrating that while they can produce useful summaries, they are occasionally prone to hallucinations~\cite{williams2024use}.
Most existing efforts have relied on large, cloud-based proprietary LLMs, such as GPT.
However, patient health information is highly sensitive, and many institutions prohibit the transmission of EHR data to external cloud services due to privacy regulations~\cite{hoffman2025medical}. 
Moreover, reliance on internet connectivity can be problematic in ambulances, rural clinics, or disaster situations where network access is limited~\cite{nissen2025medicine}. 
These challenges motivate the exploration of small language models (SLMs) that can operate locally on edge devices, bringing AI capabilities directly to the point of care~\cite{belcak2025small}.

Edge deployment of SLMs offers several advantages for healthcare settings: 
1) \textbf{Privacy}: all data is processed on-site, greatly reducing the risk of breaches or violations of laws and regulations;
2) \textbf{Low latency and mobility}: on-device inference eliminates network delays and continues to work even in offline environments such as rural zones and remote areas; and 
3) \textbf{Cost-effectiveness}: modern edge platforms, such as NVIDIA Jetson, are affordable and consume low power, enabling wide use at scale. 
These factors make on-premise AI attractive for time-critical clinical applications~\cite{basit2024medaide, nissen2025medicine}. 
Recent quantized SLMs in the 2–7 billion parameter range can now fit within 4–8GB of RAM in real-world deployments with reasonable inference time~\cite{li2025pushing, swaminathan2025benchmarking}. 
This marks a shift: fully useful LLM functionality is increasingly achievable without reliance on cloud‑scale infrastructure.
Additionally, recent surveys highlight SLMs as a new paradigm for healthcare AI at the edge \cite{garg2025rise, zheng2025review}.

While recent works have made significant progress in LLM applications for emergency medicine, several gaps remain that motivate our work.
First, despite exploration of edge deployment of medical LLMs, no prior work has tackled on-device EHR summarization for emergency physicians. Most existing clinical summarization systems rely on cloud models~\cite{ahsan2024retrieving}, which raise privacy concerns and connectivity risks in critical care settings. 
Second, existing approaches target generic clinical summaries or inpatient handoff notes~\cite{hartman2024developing}. 
However, emergency physicians have unique information needs under time pressure, requiring immediate access to critical facts alongside context-specific information about the patient’s complaints.
Third, SLMs have been deployed for general medical Q\&A or decision support~\cite{garg2025rise}, but not ED summarization generation. To the best of our knowledge, no attempt has been made to develop an EHR summarization system with limited memory and computational resources, and the corresponding challenges associated with running such a system on edge devices remain unknown. 
Lastly, automated evaluations suitable for ED summaries have not been well developed due to the lack of a gold standard summary, which makes traditional metrics obsolete.

In this paper, we address these gaps by presenting the first fully offline, edge-resident EHR summarization system tailored to emergency medicine. 
Our approach introduces a dual-stage, lightweight patient chart summarization system that maximizes the utility of resource-constrained hardware and incorporates automated quality assurance to ensure the generated summaries meet clinical standards.
By processing data locally on inexpensive IoT devices, we inherently preserve patient privacy and reduce dependency on internet connectivity. 
In our summarization system, we use two NVIDIA Jetson Orin Nano boards working in tandem, each handling a different stage of the summarization process. 
In the first stage, the first Nano (Nano-R) retrieves and prepares relevant patient information from the EHR. In the second stage, a summarization LLM on the second Nano (Nano-S) generates the summary from the prepared context. 
By preloading the retrieval model on one device and the summarization model on the other, we avoid the overhead of swapping models in and out of a single device’s memory.
The dual devices approach also substantially reduces the processing time compared to a single-device approach. 

Beyond the system architecture, we collected feedback from emergency physicians that indicated a need for targeted summaries that present information tailored for the ED. 
We introduce a two-part summarization output: a \textit{critical summary} and a \textit{context-specific summary}. 
The critical information summary is what physicians described as “need-to-know” contents, and concisely lists the patient’s most important facts that are universally relevant regardless of the patient’s presenting complaint. 
The context-specific summary is tailored to the patient’s chief complaint. 
It highlights information from the chart that is relevant to the current context. 
By dividing the summary into these two parts, we ensure that the physicians can quickly see both the general must-know background and the pertinent details for the current issue. 
This structured approach aligns with how clinicians think in emergencies: first identify any red-flag information, then assess details relevant to the present complaint.

A further challenge we address is ensuring the quality and reliability of the summaries. 
In clinical settings, a summary that contains hallucinated or incorrect information could be dangerous \cite{williams2025evaluating}. 
Traditional summary evaluation metrics like ROUGE~\cite{lin2004rouge}, BLEU~\cite{papineni2002bleu}, or BERTScore~\cite{zhang2020bertscore} are ill-suited for ED chart summarization for three reasons: 
(i) source EHRs vary in quality across authors and settings; (ii) there is no single gold summary, as clinicians legitimately produce stylistically different yet valid summaries; and (iii) lexical overlap need not reflect clinical usefulness. 
We introduce a custom evaluation called the Factual Accuracy (FA) score to prioritize factual faithfulness to the EHR and task-oriented utility rather than word overlap with a reference. 
Our FA score uses a reverse verification process where each fact in the generated summary is checked against the original EHR content to determine if it is supported. 
We leverage the LLM-as-judge~\cite{croxford2025automating} framework to evaluate the summary in the context of the source EHR. 
We categorize each statement as \textit{Supported}, \textit{Contradicted}, or \textit{Not Found}. 
The FA score is then computed as the proportion of summary statements that are supported by the record, penalizing any contradictions or unsupported content. 
We also assess two other important dimensions of summary quality: completeness and clarity. Completeness reflects how well the summary covers the key information from the patient’s chart, and clarity reflects how easy the summary is to read and understand. 

In summary, this paper presents a novel IoT-based approach to clinical text summarization that marries the strengths of edge computing with the power of SLMs, tailored specifically for ED physicians. The key contributions of our work include:
\begin{itemize}
    \item We design a novel dual-device architecture where one embedded device handles information retrieval from source EHR data and a second device generates the summary. This division enables real-time performance on low-power hardware and keeps all data on-premises for privacy preservation.
    \item We propose a specialized summarization output format for ED use, comprising a critical information summary and a context-specific summary. This meets emergency physicians’ needs more effectively than a generic summary by providing both need-to-know and case-specific details in a structured manner.
    \item We proposed FA, a \textbf{risk-weighted, evidence-linked factuality metric} to automatically verify each fact in the summary against the original EHR data using the LLM-as-Judge framework. Our evaluation methods do not require the use of reference summaries while providing a robust measure of factual correctness of the summary.
    \item We implement our system on affordable IoT hardware (Jetson Orin Nanos) and demonstrate its feasibility in practice. Through experiments with real clinical data, we show that our approach can achieve accurate, complete, and clear summaries with significantly reduced latency. This suggests our solution is suitable for field deployment in the emergency department.
\end{itemize}

\section{Related Work}
\subsection{Summarization of ED Notes Using LLMs}
The advent of large language models (LLMs) has opened new possibilities in emergency departments (ED), such as assessing clinical acuity and emergency triage~\cite{williams2024use}. 
Recent work has explored using LLMs to generate and summarize ED documentation, aiming to reduce the burden of manual note-taking. 
Hartman et al.~\cite{hartman2024developing} developed an LLM-based system to produce ED-to-inpatient handoff summaries and evaluated its output against actual physician-written notes. 
Notably, the AI summaries introduced no major safety issues—only minor omissions or less polished phrasing made them marginally inferior to expert-written notes~\cite{hartman2024developing}. 
This suggests that LLM-based summarization can achieve high accuracy and usefulness, though a human-in-the-loop is still advisable to ensure nuanced details are captured.
Emerging tools are already leveraging AI to streamline ED documentation. 
For instance, the University of Alberta developed an AI scribe tool named "Jenkins" that transcribes clinician and patient conversations in real time, takes notes, and summarizes interactions \footnote{https://www.ualberta.ca/en/folio/2024/10/ai-scribe-could-help-emergency-docs-improve-care.html}. 
Clearly, LLMs have the potential to reduce administrative workload as well as improve the level of detail in records. 

Beyond the ED-specific setting, a range of approaches have been investigated for summarizing clinical text. 
Early efforts by Pivovarov et al. applied traditional NLP techniques to automatically summarize information from EHRs~\cite{pivovarov2015automated}. 
With the rise of LLMs, Van Veen et al.~\cite{van2023clinical,van2024adapted} demonstrated that adapted LLMs can even outperform medical experts on summarizing clinical notes while achieving higher relevance and accuracy scores than human-written summaries. 
Pal et al. similarly explored a neural approach to summarize lengthy EHR notes, showing the feasibility of LLM-based summarization in medical contexts~\cite{pal2023neural}. 
Madzime et al. proposed further enhancements to LLM summarization of EHR text to improve the coverage and usefulness of the outputs~\cite{madzime2024enhanced}. 
To facilitate evaluation of such systems, Croxford et al. introduced a standardized instrument for rating the quality of provider note summaries, which helps enable comparisons between human and LLM-generated summaries~\cite{croxford2025development}.
Notably, SLMs have proven capable of performing summarization tasks, suggesting that compact models can still perform effectively with proper optimization \cite{fu2024tiny, xu2025evaluating}. 
Xu et al. evaluated the summarization abilities of various SLMs on news articles and found that, with fine-tuning, compact models could achieve results close to those of much larger models~\cite{xu2025evaluating}. 
Likewise, Fu et al. examined meeting transcript summarization and observed that even SLMs can outperform larger LLMs given appropriate training strategies and prompts~\cite{fu2024tiny}. 
Moreover, Hemamou et al. leverage LLMs to extract summaries of lengthy texts, effectively scaling summarization to inputs beyond a single model’s usual context window~\cite{hemamou2024scaling}. 


\begin{figure}[t]
  \centering
  \includegraphics[width=0.49\textwidth]{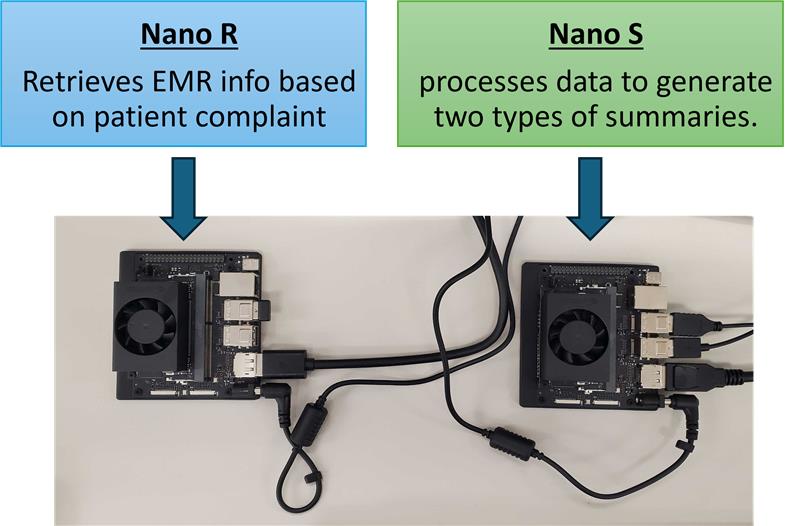}
  \caption{Picture of our two Jetson Nanos connected locally, Nano-R (left) retrieves EHR context, while Nano-S (right) generates the two-mode summary (critical + context-specific) on-device.}
  \label{fig:real_nano}
\end{figure}

\subsection{LLMs for Clinical Decision Support in Emergency Medicine}

Beyond documentation, large language models are being studied as decision support tools in the ED. 
Williams et al. showed that GPT-4 could accurately determine which of two ED patients was more critically ill by reading their initial notes~\cite{williams2024use}. 
Arvidsson et al. pitted GPT-4 against experienced doctors on complex primary care cases and found the AI lagging across multiple dimensions~\cite{arvidsson2024chatgpt}. 
Similarly, Hager et al.~\cite{hager2024evaluation} conducted a comprehensive evaluation of state-of-the-art LLMs on 2,400 real hospital cases and found that even fine-tuned models fell far short of physician accuracy in full clinical decision-making tasks. 

\subsection{On-Device LLMs and Edge Computing in Healthcare}
An important emerging trend is deploying language models on edge devices. 
These SLMs are designed to run under constrained computing resources and are increasingly viewed as a way to bring AI assistance directly to the point of care~\cite{qu2025mobile,zheng2025review}. 
By combining edge computing with lightweight models, health professionals can get real-time AI support on-site and even in environments with limited internet connectivity or stringent data privacy requirements~\cite{nissen2025medicine}. 
Basit et al. introduced MedAide, an on-premise medical chatbot that runs on a small NVIDIA Jetson edge device by leveraging a fine-tuned 7B-parameter model optimized via LoRA~\cite{basit2024medaide}. 
Another study by Nissen et al. benchmarked several such compact models on clinical reasoning tasks~\cite{nissen2025medicine}. 
They found that a general-purpose 2.7B-parameter model (Phi-3 Mini) offered an excellent speed to accuracy tradeoff on mobile-grade hardware, while slightly larger medically fine-tuned models reached the highest diagnostic accuracy at the cost of slower inference~\cite{nissen2025medicine}. 
Recent surveys cover techniques such as quantization, distillation, and efficient model architectures that enable large models to run in resource-constrained environments~\cite{qu2025mobile,zheng2025review}. 
Real-world use cases for edge AI in emergency care are beginning to emerge. NTT Data researchers, for instance, have prototyped an AI integrated into a portable ultrasound machine, and wearable monitors with built-in AI can continuously watch a patient’s vital signs and alert clinicians to dangerous trends~\cite{anusha2024emerging}.

\begin{figure*}[t]
  \centering
\includegraphics[width=0.98\textwidth]{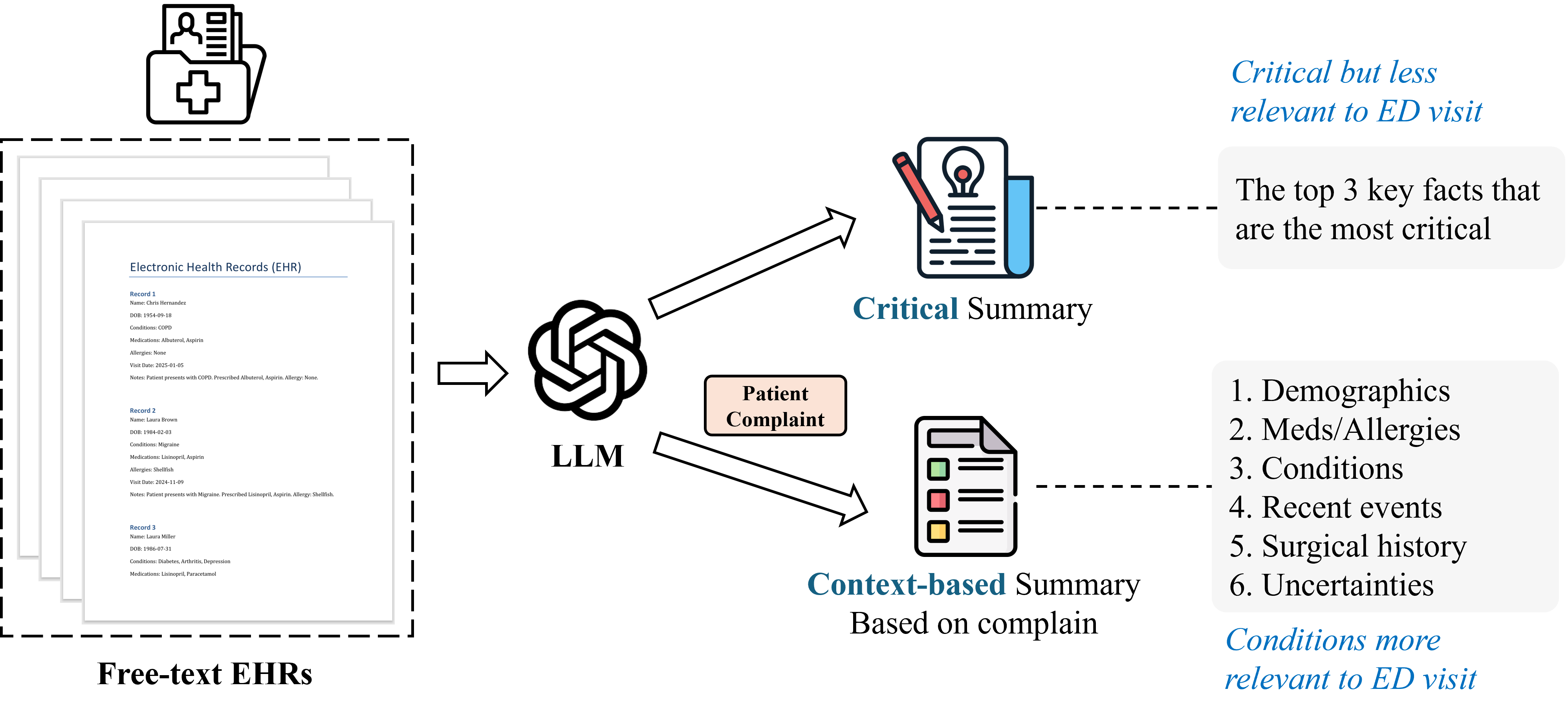}
  \caption{Two-mode ED chart summarization: from EHR input and the chief complaint, the system produces (i) a \textit{Critical} summary of the top must-know facts and (ii) a \textit{Context-based} summary focused on the current complaint.}
  \label{fig:cycle}
\end{figure*}

\section{Methodology}

\subsection{System Overview and Edge Hardware}

Our summarization system is implemented across two IoT devices, which we refer to as \textbf{Nano-R} (for Retrieval) and \textbf{Nano-S} (for Summarization). 
Each device is an NVIDIA Jetson Orin Nano with a CUDA-enabled GPU and 8GB RAM. 
These devices are inexpensive and energy-efficient, making them suitable for point-of-care deployment. 
Figure \ref{fig:real_nano} demonstrates our setup: Nano-R hosts the patient data and performs retrieval, then transmits the relevant text to Nano-S over a Python socket. 
Nano-S runs the SLM to generate the summary, which is then presented to the clinician. 
Both stages run completely offline, where no Internet is required after the models and data have been initially set up.

Splitting the workload between two devices addresses the major constraint of edge deployment, which is limited memory and compute for running LLMs. 
Loading even a 7B-parameter model in 4-bit precision typically requires about 6-7GB of GPU memory, leaving little room for additional context or processes on our 8GB device. 
By dedicating Nano-S exclusively to the summarization model, we avoid competition with data indexing tasks. Our dual setup also significantly reduces wait times when running multiple summaries with the same LLM in series. 
Since the Nano S does not require retrieval, it can simply keep the LLM on its RAM and does not need to reload it for every subsequent summary.
Meanwhile, Nano-R can handle retrieval computations in parallel, potentially preparing the relevant text even while Nano-S is still loading the model or finishing a prior request. 
This pipelined concurrency hides some of the latency and enables a quicker turnaround for successive queries.

\subsection{Stage 1: Retrieval Module (Nano-R)}

The retrieval stage is responsible for extracting from the patient’s EHR the content most pertinent to the clinician’s query or current information need.

\subsubsection{Data Processing} 
Our parser scans each note for known section headers and divides the text using double newline patterns into meaningful chunks. 
We apply a simpler heuristic by splitting the text by paragraph boundaries and inserting breaks at a fixed token length if no clear paragraph separators exist. 
After splitting the notes, each section/chunk is encoded into a vector embedding using the \textit{BGE-M3 embedding model}. 
All embeddings for a patient’s record are stored in a local vector index. 
When the emergency physician enters a complaint (for example: "chest pain"), the retrieval module encodes this query with the embedding model to obtain a query embedding.
It then performs a FAISS search against the patient’s index to retrieve the top-$k$ most relevant sections. 
In our setup, we set $k=5$, meaning the system will pull the five sections of the EHR most likely to contain information related to the query. 
These top-$k$ sections (with their original text) are then packaged as the \textit{retrieved context} for the next stage.

\subsubsection{Design Rationale} 
This retrieval step serves to drastically narrow down the input for the summarization SLMs. Rather than feeding an LLM the patient’s entire record, we provide only the most relevant parts. 

\subsection{Stage 2: Summarization Module (Nano-S)}
The summarization stage runs on the Nano-S device and is powered by a local SLM that generates the final summary text. 
Upon receiving the retrieved context from Nano-R, the Nano-S process constructs a prompt for the model that includes: 
1) the retrieved EHR sections as context and 
2) description of the task and desired output format as a prompt. 

\begin{table*}[t]
\centering
\begin{tabular}{ccc|ccc|ccc|cc}
\hline\hline
\multirow{2}{*}{Dataset} & \multirow{2}{*}{Prompt Technique} & \multirow{2}{*}{LLM} & 
\multicolumn{3}{c|}{Critical Summary} & 
\multicolumn{3}{c|}{Context Summary} & 
\multicolumn{2}{c}{Average} \\
&&& FA (CSR$|$CR$|$UR) & CO & CL & FA (CSR$|$CR$|$UR) & CO & CL & Crit & Ctx \\
\hline
\multirow{27}{*}{MIMIC} 
 & \multirow{6}{*}{Zero-Shot} 
 & Starling-lm & \fabox{0.46}{0.32}{0.10}{0.58} & 2.00 & 3.00 & \fabox{4.58}{0.95}{0.05}{0.00} & 3.00 & 4.00 & 1.82 & 3.86 \\
 &  & Phi3 & \fabox{2.19}{0.62}{0.19}{0.19} & 2.00 & 2.00 & \fabox{3.12}{0.75}{0.12}{0.12} & 3.00 & 3.00 & 2.06 & 3.04 \\
 &  & Neural-Chat & \fabox{3.44}{0.75}{0.00}{0.25} & 3.00 & 4.00 & \fabox{3.44}{0.75}{0.00}{0.25} & 4.00 & 4.00 & 3.48 & 3.81 \\
 &  & Mistral & \fabox{5.00}{1.00}{0.00}{0.00} & 4.00 & 4.00 & \fabox{3.80}{0.85}{0.11}{0.04} & 3.00 & 3.00 & 4.33 & 3.27 \\
 &  & Zephyr & \fabox{5.00}{1.00}{0.00}{0.00} & 4.00 & 5.00 & \fabox{2.32}{0.57}{0.00}{0.43} & 4.00 & 4.00 & 4.67 & 3.44 \\
 &  & Openchat & \fabox{2.98}{0.69}{0.04}{0.27} & 3.00 & 3.00 & \fabox{5.00}{1.00}{0.00}{0.00} & 3.00 & 3.00 & 2.99 & 3.67 \\
\cline{2-11}
 & \multirow{6}{*}{Few-Shot} 
 & Starling-lm & \fabox{5.00}{1.00}{0.00}{0.00} & 5.00 & 5.00 & \fabox{3.09}{0.77}{0.18}{0.06} & 3.00 & 4.00 & 5.00 & 3.36 \\
 &  & Phi3 & \fabox{1.62}{0.50}{0.10}{0.40} & 3.00 & 3.00 & \fabox{2.32}{0.57}{0.00}{0.43} & 3.00 & 3.00 & 2.54 & 2.77 \\
 &  & Neural-Chat & \fabox{5.00}{1.00}{0.00}{0.00} & 4.00 & 5.00 & \fabox{3.75}{0.80}{0.00}{0.20} & 3.00 & 3.00 & 4.67 & 3.25 \\
 &  & Mistral & \fabox{2.50}{0.60}{0.00}{0.40} & 4.00 & 4.00 & \fabox{3.21}{0.71}{0.00}{0.29} & 3.00 & 4.00 & 3.50 & 3.40 \\
 &  & Zephyr & \fabox{2.13}{0.63}{0.22}{0.15} & 3.00 & 3.00 & \fabox{3.93}{0.86}{0.07}{0.07} & 4.00 & 4.00 & 2.71 & 3.98 \\
 &  & Openchat & \fabox{4.22}{0.88}{0.00}{0.12} & 4.00 & 4.00 & \fabox{3.33}{0.73}{0.00}{0.27} & 2.00 & 3.00 & 4.07 & 2.78 \\
\cline{2-11}
 & \multirow{6}{*}{CoT} 
 & Starling-lm & \fabox{5.00}{1.00}{0.00}{0.00} & 4.00 & 4.00 & \fabox{2.85}{0.71}{0.14}{0.14} & 3.00 & 4.00 & 4.33 & 3.29 \\
 &  & Phi3 & \fabox{2.36}{0.65}{0.18}{0.18} & 3.00 & 4.00 & \fabox{1.79}{0.56}{0.19}{0.25} & 3.00 & 3.00 & 3.12 & 2.60 \\
 &  & Neural-Chat & \fabox{1.98}{0.58}{0.17}{0.25} & 4.00 & 5.00 & \fabox{5.00}{1.00}{0.00}{0.00} & 3.00 & 4.00 & 3.66 & 4.00 \\
 &  & Mistral & \fabox{5.00}{1.00}{0.00}{0.00} & 3.00 & 4.00 & \fabox{3.93}{0.86}{0.07}{0.07} & 3.00 & 4.00 & 4.00 & 3.64 \\
 &  & Zephyr & \fabox{4.38}{0.90}{0.00}{0.10} & 4.00 & 4.00 & \fabox{2.67}{0.66}{0.07}{0.28} & 4.00 & 4.00 & 4.12 & 3.56 \\
 &  & Openchat & \fabox{5.00}{1.00}{0.00}{0.00} & 4.00 & 4.00 & \fabox{4.48}{0.94}{0.06}{0.00} & 3.00 & 4.00 & 4.33 & 3.83 \\
\cline{2-11}
 & \multirow{6}{*}{Self-Ask} 
 & Starling-lm & \fabox{3.19}{0.75}{0.10}{0.15} & 4.00 & 4.00 & \fabox{5.00}{1.00}{0.00}{0.00} & 3.00 & 4.00 & 3.73 & 4.00 \\
 &  & Phi3 & \fabox{0.99}{0.42}{0.16}{0.42} & 3.00 & 4.00 & \fabox{0.00}{0.19}{0.12}{0.69} & 3.00 & 3.00 & 2.66 & 2.00 \\
 &  & Neural-Chat & \fabox{0.18}{0.29}{0.14}{0.57} & 3.00 & 3.00 & \fabox{5.00}{1.00}{0.00}{0.00} & 3.00 & 4.00 & 2.06 & 4.00 \\
 &  & Mistral & \fabox{5.00}{1.00}{0.00}{0.00} & 4.00 & 4.00 & \fabox{5.00}{1.00}{0.00}{0.00} & 3.00 & 4.00 & 4.33 & 4.00 \\
 &  & Zephyr & \fabox{2.92}{0.67}{0.00}{0.33} & 4.00 & 4.00 & \fabox{1.71}{0.50}{0.07}{0.43} & 4.00 & 4.00 & 3.64 & 3.24 \\
 &  & Openchat & \fabox{5.00}{1.00}{0.00}{0.00} & 4.00 & 5.00 & \fabox{3.91}{0.88}{0.12}{0.00} & 3.00 & 4.00 & 4.67 & 3.64 \\
\hline\hline
\end{tabular}
\vspace{4pt}
\caption{Evaluation with per-summary averages. FA cells show S$|$C$|$U rates (CSR$|$CR$|$UR). Two averages are reported: Avg Crit = mean(fa,comp,clar) for Critical; Avg Ctx for Context.}
\label{tab:llm_eval_MIMIC}
\end{table*}

\subsubsection{Summarization Output Format}
We instruct the SLMs on Nano-s to produce two different kinds of summaries:
\begin{itemize}
    \item \textbf{Critical Findings:} 
    A bulleted list of three items that the physician should know about this patient, such as continuing or recurring issues.
    \item \textbf{Context-Specific Summary:} 
    A paragraph including demographics, meds/allergies, conditions, recent events, surgical history, uncertainties, relevant to the patient’s complaints.
\end{itemize}
We design all prompts to elicit the same two-part summarization output. 
Beyond this format, we tested multiple versions of the prompt using several widely adopted prompting techniques to obtain more comprehensive results across models.

\subsubsection{Local LLM Selection}
Running an LLM on a Jetson Nano imposes strict limits on model size and architecture. 
Our goal was to choose a model that maximizes summarization quality while staying within the device’s capabilities. 
We evaluated \textbf{12 candidate models} ranging from about 3 billion to 7 billion parameters.
Given these observations, we selected the appropriate models to run in our experiment section.

\begin{table*}[t]
\centering
\begin{tabular}{ccc|ccc|ccc|cc}
\hline\hline
\multirow{2}{*}{Dataset} & \multirow{2}{*}{Prompt Technique} & \multirow{2}{*}{LLM} & 
\multicolumn{3}{c|}{Critical Summary} & 
\multicolumn{3}{c|}{Context Summary} & 
\multicolumn{2}{c}{Average} \\
&&& FA (CSR$|$CR$|$UR) & CO & CL & FA (CSR$|$CR$|$UR) & CO & CL & Crit & Ctx \\
\hline
\multirow{20}{*}{Real} & \multirow{4}{*}{Zero-Shot} 
& Starling-lm & \fabox{3.74}{0.81}{0.03}{0.16} & 2.40 & 4.00 & \fabox{4.68}{0.96}{0.04}{0.00} & 1.80 & 4.00 & 3.38 & 3.49 \\
 &  & Neural-Chat & \fabox{4.57}{0.94}{0.03}{0.03} & 2.60 & 4.60 & \fabox{4.30}{0.90}{0.03}{0.08} & 2.00 & 4.00 & 3.92 & 3.43 \\
 &  & Mistral & \fabox{3.55}{0.80}{0.08}{0.12} & 2.80 & 4.60 & \fabox{4.50}{0.93}{0.02}{0.05} & 2.00 & 3.80 & 3.65 & 3.43 \\
 &  & Openchat & \fabox{4.88}{0.98}{0.00}{0.02} & 3.60 & 5.00 & \fabox{4.54}{0.93}{0.01}{0.06} & 2.00 & 4.00 & 4.49 & 3.51 \\
\cline{2-11}
 & \multirow{4}{*}{Few-Shot} 
 & Starling-lm & \fabox{4.37}{0.92}{0.04}{0.04} & 3.40 & 4.60 & \fabox{4.23}{0.88}{0.01}{0.11} & 2.00 & 3.80 & 4.12 & 3.34 \\
 &  & Neural-Chat & \fabox{3.68}{0.81}{0.05}{0.14} & 4.40 & 5.00 & \fabox{3.82}{0.82}{0.01}{0.17} & 2.00 & 4.20 & 4.36 & 3.34 \\
 &  & Mistral & \fabox{3.66}{0.80}{0.03}{0.18} & 4.40 & 4.80 & \fabox{1.70}{0.52}{0.12}{0.36} & 1.80 & 3.60 & 4.29 & 2.37 \\
 &  & Openchat & \fabox{4.30}{0.92}{0.08}{0.00} & 3.40 & 4.40 & \fabox{4.56}{0.93}{0.01}{0.06} & 2.00 & 3.80 & 4.03 & 3.45 \\
\cline{2-11}
 & \multirow{4}{*}{CoT} 
 & Starling-lm & \fabox{3.45}{0.80}{0.12}{0.08} & 4.40 & 4.80 & \fabox{4.62}{0.94}{0.01}{0.04} & 2.00 & 4.00 & 4.22 & 3.54 \\
 &  & Neural-Chat & \fabox{4.38}{0.90}{0.00}{0.10} & 5.00 & 5.00 & \fabox{4.72}{0.96}{0.01}{0.03} & 2.00 & 3.80 & 4.79 & 3.51 \\
 &  & Mistral & \fabox{4.56}{0.95}{0.05}{0.00} & 3.80 & 4.80 & \fabox{4.18}{0.88}{0.02}{0.10} & 2.00 & 3.80 & 4.39 & 3.33 \\
 &  & Openchat & \fabox{3.99}{0.87}{0.09}{0.04} & 3.40 & 5.00 & \fabox{4.39}{0.92}{0.04}{0.04} & 2.00 & 4.00 & 4.13 & 3.46 \\
\cline{2-11}
 & \multirow{4}{*}{Self-Ask} 
 & Starling-lm & \fabox{3.85}{0.85}{0.09}{0.06} & 2.60 & 4.00 & \fabox{4.37}{0.92}{0.05}{0.03} & 2.00 & 4.00 & 3.48 & 3.46 \\
 &  & Neural-Chat & \fabox{4.32}{0.90}{0.02}{0.09} & 2.20 & 4.20 & \fabox{4.28}{0.90}{0.03}{0.08} & 2.00 & 4.00 & 3.57 & 3.43 \\
 &  & Mistral & \fabox{3.84}{0.84}{0.06}{0.09} & 4.40 & 4.80 & \fabox{4.96}{1.00}{0.00}{0.00} & 2.00 & 4.00 & 4.35 & 3.65 \\
 &  & Openchat & \fabox{3.58}{0.80}{0.07}{0.13} & 3.60 & 4.60 & \fabox{4.06}{0.87}{0.05}{0.08} & 2.00 & 3.80 & 3.93 & 3.29 \\
\cline{2-11}
 & \multirow{4}{*}{Plan-and-Solve} 
 & Starling-lm & \fabox{3.24}{0.75}{0.07}{0.18} & 3.40 & 4.60 & \fabox{4.68}{0.96}{0.02}{0.02} & 2.00 & 3.80 & 3.75 & 3.49 \\
 &  & Neural-Chat & \fabox{4.06}{0.87}{0.06}{0.07} & 3.40 & 4.40 & \fabox{4.53}{0.92}{0.00}{0.08} & 1.80 & 4.00 & 3.95 & 3.44 \\
 &  & Mistral & \fabox{3.77}{0.82}{0.03}{0.15} & 3.40 & 4.80 & \fabox{4.26}{0.90}{0.05}{0.05} & 2.00 & 3.80 & 3.99 & 3.35 \\
 &  & Openchat & \fabox{3.53}{0.76}{0.00}{0.24} & 3.40 & 4.60 & \fabox{4.43}{0.91}{0.00}{0.09} & 2.00 & 4.20 & 3.84 & 3.54 \\
\hline\hline
\end{tabular}
\vspace{4pt}
\caption{Evaluation with per-summary averages. FA is scaled to [0,5]; cells show FA over S$|$C$|$U (CSR$|$CR$|$UR). Averages: Crit = mean(FA, CO, CL) over Critical; Ctx = mean(FA, CO, CL) over Context.}
\label{tab:llm_eval_prompt_REAL}
\end{table*}

\subsubsection{Evaluation: LLM-as-Judge and Metrics}
In the ED setting, there are no gold-standard reference summaries, making conventional summarization metrics such as ROUGE or BERTScore inapplicable. 
Furthermore, the quality of any generated summary is fundamentally constrained by the quality of the underlying EHR; incomplete or inaccurate records inevitably lead to suboptimal summaries. 
For ED physicians, factual reliability is the most important concern, rather than stylistic quality or adherence to an idealized target. 

This motivates our use of an LLM-as-a-judge framework~\cite{croxford2025automating} to directly verify each generated claim against the source EHR. 
The aim is to detect three types of outcomes: \textit{SUPPORTED} claims, which match the source evidence with correct temporality, negation, and numeric details; \textit{CONTRADICTED} claims, which conflict with the record; and \textit{NOT\_FOUND} claims, which are unsupported by any evidence in the EHR.

Let $N$ denote the total number of atomic claims extracted from the generated summary. 
For each claim, we retrieve top-$k$ relevant snippets from the original EHR and pass the claim–evidence pair to an LLM verifier with strict JSON output. The verifier classifies each claim into one of the following categories:
\begin{itemize}
    \item \textbf{SUPPORTED}: Explicitly supported by the EHR with correct temporality, negation, and numeric values.
    \item \textbf{CONTRADICTED}: Directly conflicts with the EHR.
    \item \textbf{NOT\_FOUND}: No supporting evidence found.
\end{itemize}

We define the counts:
\begin{align}
S &= \#\{\text{SUPPORTED claims}\}, \\
C &= \#\{\text{CONTRADICTED claims}\}, \\
U &= \#\{\text{NOT\_FOUND claims}\}.
\end{align}

By construction,
\begin{equation}
N = S + C + U.
\end{equation}

The normalized proportions are then defined as:
\begin{equation}
\delta_s = \frac{S}{N},\delta_c = \frac{C}{N}, \delta_u = \frac{U}{N}
\end{equation}
with $\delta_s + \delta_c + \delta_u = 1$.

Contradicted claims carry a strong penalty and unsupported claims are neutral to reflect clinical risk.
The raw factual accuracy score is:
\begin{equation}
FA_{\text{raw}} = w_s \cdot \delta_s + w_c \cdot \delta_c + w_u \cdot \delta_u,
\end{equation}
and the clipped score is
\begin{equation}
FA = \min\{5, \max\{0, FA_{\text{raw}}\}\}.
\end{equation}
For the Risk-Weighted Scoring, we assign weights $(w_s, w_c, w_u)$ to $w_s = 5$, $w_c = -3.75$, and $w_u = 1.25$.

Furthermore, we use the LLM-as-a-judge framework to assess the completeness and clarity of the summary, where: 
\begin{itemize}
    \item \textbf{Completeness:} Does the summary include all the critical information from the source that is relevant to the query and important for decision-making?
    \item \textbf{Clarity:} Is the summary concise, coherent, and easy to interpret in a clinical setting?
\end{itemize}

\section{Experiments}

\subsection{Hardware Configuration}
\subsubsection{Edge Devices} 2 Jetson Orin Nano 8 GBs are used for both the Jetson Nano-R and Nano-S. The Jetson Nano developer kit features a 6-core Arm Cortex-A78AE v8.2 64-bit CPU and a 1024-core NVIDIA Ampere architecture GPU with 32 Tensor Cores. It includes 8GB of LPDDR5 memory.

\subsubsection{System} Our system runs on Ubuntu 22.04 with L4T version 36.4.4, corresponding to JetPack 6.0. 
Model inference is performed using Ollama, and retrieval is handled using FAISS and Python libraries such as socket and numpy. 

\subsubsection{Communication} A basic Python socket connection is used, with the Jetson Nano-S acting as the EHRs summarizer and Jetson Nano-R as the EHRs retrieval. 
Data is sent over a TCP/IP link using default socket settings.

\subsubsection{Optimizations} All SLMs quantized to 4-bit precision using Ollama's built-in quantization support.
No additional pruning or distillation techniques are applied.

\subsection{Software Configuration}
\subsubsection{Datasets} 
We evaluate our dual-stage summarizer on 2 datasets, MIMIC-IV-Note: Deidentified free-text clinical notes~\cite{johnson2023mimicivnote} and real-world EHRs collected from Rockyview General Hospital. For context-specific summaries, we used the complaint "chest pain" as the context.

\subsubsection{Models} 
In this study, we tested several SLMs ranging from 3.8B to 7B parameters.
After initial testing, 6 SLMs are chosen, including Phi-3-mini, Starling-LM, Neural-Chat, Mistral, Zephyr, and OpenChat, to proceed with the Mimic experiment.
Following the Mimic experiment, we selected the top four performing SLMs: Starling-LM, Neural-Chat, Mistral, and OpenChat, for the experiment using real data.

\subsubsection{Prompting techniques} 
We employed 5 existing prompting techniques, including zero-shot, few-shot, Chain-of-Thought (CoT)~\cite{wei2022chain}, Self-Ask~\cite{press2022measuring}, and Plan-and-Solve~\cite{wang2023plan}.

\subsubsection{Evaluation}
For MIMIC evaluation, we deployed GPT-4o in a secure cloud environment via Azure AI Foundry to ensure compliance with MIMIC data-use restrictions. 
For real-world EHRs evaluation, we deployed GPT-oss-20b model locally on Ollama for fully offline evaluation.

\subsection{Results}
\subsubsection{Model Selection on MIMIC-IV Notes}
We first evaluated critical and context summarization on a range of open-source LLMs on the MIMIC-IV-Note to identify the best candidates for on-device summarization. 
Our focus was primarily on 7B-parameter chat models with strong general performance.
Each model was tested under five prompting strategies, from classic to more recent, to measure our proposed FA score, which determines how well the summary’s facts aligned with the source EHR. 
On MIMIC, Starling-LM, Neural-Chat, Mistral, and OpenChat repeatedly hit $FA = 5.0$ in at least one setting, with high CO and CL scores. 
Starling-LM produced the most factually complete summaries, with significantly fewer CR and UR. It had one of the highest supported-fact rates (CSR) under multiple prompts.
Neural-Chat closely matched Starling’s performance, yielding high CSR and low CR\&UR. 
OpenChat and Mistral-7B also performed strongly compare with Phi3 and Zephyr.
In summary, these four models were chosen because they delivered the highest factual accuracy on MIMIC, each model’s CR was near zero in at least one prompting strategy, per Table~\ref{tab:llm_eval_MIMIC}.

\subsubsection{Evaluation on Real Patient Dataset} 
Next, we conducted a comprehensive evaluation of real-world EHRs to compare the four chosen models. 
We applied the same prompting techniques on all models with an additional \textit{plan-and-solve} to add more baselines. 
Table~\ref{tab:llm_eval_prompt_REAL} presents the summarization results for each model. 

\textbf{Zero-Shot is the strongest overall prompt on-device:}
Averaged across the four models, Zero-Shot attains the highest overall FA at 4.35 and the highest mean Context FA 4.51. 
CoT is second overall at 4.29, followed by Self-Ask at 4.16 and Plan-and-Solve at 4.06; Few-Shot is last at 3.79.
On an 8 GB Nano, prompt tokens compete with retrieved context. 
We find that shorter, zero-shot prompts preserve more EHR evidence within the context window, lowering UR without a corresponding rise in CR. 
On SLMs, heavier few-shot prompts further depress FA by introducing distractions and exemplar-anchoring effects that reduce effective evidence retrieval.

\textbf{Best model and prompt pairings:}
OpenChat + Zero-Shot achieved the best overall score at 4.71, combining critical and context FA scores. 
Neural-Chat + CoT followed at 4.55 as the top reasoning-oriented setup, yielding the highest Crit Avg at 4.79 with perfect CO and CL score. 
Mistral + Self-Ask reached 4.40 overall, excelling on the context summary and delivering near-perfect FA at 4.96. 
From a deployment perspective, when under time pressure, choose OpenChat + Zero-Shot; when explicit reasoning is desired, use Neural-Chat + CoT; and when richer contextual detail is the priority, prefer Mistral + Self-Ask.

\textbf{Model vs. Prompt:}
We quantify FA variation attributable to prompts vs. models. 
Aggregating both summaries, FA ranges 3.79 to 4.35 across prompts (spread = 0.56) versus 3.90 to 4.27 across models (spread = 0.37). 
By summary type, Critical FA prompt spread is 0.53 compared to model spread of 0.47, whereas Context FA prompt spread is 0.93 compared with 0.60. 
In our experiment, prompt choice accounts for more FA variance than model choice, particularly for context summary.
Across models and prompts, CR is near-zero on real EHRs, and FA losses come mostly from UR, which further favors shorter prompts that preserve more retrieved text in context.

\subsubsection{Speed} 
End-to-end summarization took between a few seconds and a minute, depending on the model size. 
For instance, using a smaller 2.7B model, the pipeline generated a summary in around 40 seconds, whereas a larger 3.8B model took about 70 seconds on the same patient data. 
In addition, we found that prompting techniques can affect performance. For example, using a few-shot prompt improved summary conciseness and reduced generation time, whereas using a CoT prompt increased detail at some cost to speed. 

\subsubsection{Ablation Study}
We performed an ablation study to compare running our summarization system on a single Jetson Nano versus our dual setup. 
The single-device configuration processes both retrieval and summarization sequentially on one Nano, whereas our dual setup offloads retrieval to Nano-R and runs the LLM summarization on Nano-S. 
Our separation shows a notable performance gain shown in Table~\ref{tab:nano_runtime_breakdown}. 
In our tests, using two Nanos nearly halved the end-to-end summarization time compared to a single Nano. 
Equally important, the dual configuration improved system stability and throughput. On a single Nano, memory contention forced the LLM to reload from scratch for each query and limited us to smaller models.
We observed that running a 7B model end-to-end on one 8 GB Nano often pushed the device to its limits or even caused crashes, as was the case with Llama2-7B. 
In the two-device setup, Nano-S can keep the large model in memory between runs, avoiding repetitive loading. 
The dual-Jetson pipeline consistently outperformed the single-Jetson deployment in latency. 
In summary, we confirm that our dual setup can reduce latency via parallel execution and model persistence, and increase capacity by allowing larger models to run reliably. 


\begin{table}[tb]
\centering
\small  
\setlength{\tabcolsep}{4pt}  
\begin{tabular}{lccc}
\hline\hline
\textbf{Stage} & \textbf{Single} & \multicolumn{2}{c}{\textbf{Dual-Nano}} \\
\cline{3-4}
 & \textbf{Nano} & First Run & Subsequent Run \\
\hline
Model load               & 58.79  & 52.17  & 0.02 \\
Retrieval gen.           & 28.82  & 0.00   & 0.00 \\
Summarization gen.       & 160.36 & 155.08 & 153.72 \\
\hline
\textbf{Total}           & 247.97 & 207.25 & 153.74 \\
\hline\hline
\end{tabular}
\caption{Runtimes by stage on a single vs. dual Jetson Nano. First dual run includes model initialization overhead.}
\label{tab:nano_runtime_breakdown}
\end{table}

\section{Future Work}
In future work, we will collaborate with practicing emergency physicians to rate summaries for their accuracy, completeness, and clarity.
We would validate our FA metric by comparing model-assigned scores to clinician adjudications and having clinicians check the FA directly on the source EHRs.
On our dual-device system, we will work on fine-tuning SLMs for edge inference in order to improve the quality.
Eventually, we will develop customized prompting strategies that are tailored to the ED and our dual-device system.

\section{Conclusion}
We have presented a dual-stage edge solution for summarizing health records in emergency care, which shows that SLMs deployed on low-power IoT devices without cloud support can achieve excellent performance. 
Our Edge-Resident Two-Stage EHR Summarizer combines retrieval-augmented generation, dual-device deployment, and customized prompts to meet the demanding requirements of emergency medicine. 
Our approach balances the limitations of SLMs by combining them efficiently: the retrieval stage lightens the load on the generation stage, and the two-part output ensures that no critical information is missed even in a concise summary. 
Finally, we present a new method for evaluating the quality of EHR summarization based on evidence without gold references.

\bibliographystyle{IEEEtran}
\bibliography{references}
\end{document}